\documentclass[10pt,twocolumn,letterpaper]{article}

\usepackage[pagenumbers]{cvpr} 

\usepackage{graphicx}
\usepackage{amsmath}
\usepackage{amssymb}
\usepackage{booktabs}
\usepackage{subfiles}

\usepackage[pagebackref,breaklinks,colorlinks]{hyperref}

\usepackage[capitalize]{cleveref}
\crefname{section}{Sec.}{Secs.}
\Crefname{section}{Section}{Sections}
\Crefname{table}{Table}{Tables}
\crefname{table}{Tab.}{Tabs.}

\def\cvprPaperID{20} 
\def\confName{CVPR}
\def\confYear{2022}

\begin{document}

\title{ET tu, CLIP? Addressing Common Object Errors for Unseen Environments}

\author{Ye Won Byun\thanks{\hspace{4pt}Everyone Contributed Equally -- Alphabetical order}
\and
Cathy Jiao$^*$
\and
Shahriar Noroozizadeh$^*$
\and
Jimin Sun$^*$
\and
Rosa Vitiello$^*$
\vspace{2pt}
\and
Carnegie Mellon University \\
{\tt\small \{yewonb,cljiao,snoroozi,jimins2,rvitiell\}@andrew.cmu.edu}
}
\maketitle

\begin{abstract}
We introduce a simple method that employs pre-trained CLIP encoders to enhance model generalization in the ALFRED task. In contrast to previous literature where CLIP replaces the visual encoder, we suggest using CLIP as an additional module through an auxiliary object detection objective. We validate our method on the recently proposed Episodic Transformer architecture and demonstrate that incorporating CLIP improves task performance on the unseen validation set. Additionally, our analysis results support that CLIP especially helps with leveraging object descriptions, detecting small objects, and interpreting rare words.
\end{abstract}


\section{Introduction}
Embodied instruction following (EIF) tasks entail executing fine-grained navigation and interaction action sequences according to natural language directives. This requires processing and understanding information from heterogeneous sources to successfully navigate and interact with unseen environments \cite{min2021film, blukis2022persistent}. 
In the multimodal research community, large-scale pre-trained models have been shown to improve multimodal alignment and generalization performance \cite{li2021supervision,thomason2022language, parisi2022unsurprising, cui2022democratizing}. In particular, several recent works evaluate the CLIP (Contrastive Language Image Pretraining) \cite{pmlr-v139-radford21a} model's capabilities for embodied AI tasks, including object navigation \cite{https://doi.org/10.48550/arxiv.2203.10421, khandelwal2021simple} and vision language navigation \cite{DBLP:journals/corr/abs-2107-06383,tam2022semantic, liang2022visual}. The most common approach in this direction has been simply replacing the visual encoder with CLIP's visual encoder.

In this work, we hypothesize that pre-training on large-scale image-text pairs will induce more generalizable multimodal representations, leading to better performance in unseen environments of the ALFRED task \cite{shridhar2020alfred}. In contrast to previous literature, we propose a simple model-agnostic method to use CLIP as an auxiliary module to take advantage of CLIP's multimodal alignment capabilities. Concretely, we introduce a novel object detection loss without having to change the model's architecture. 
We investigate the proposed method through preliminary experiments based on the Episodic Transformer (ET) \cite{pashevich2021episodic} architecture, a competitive system on the ALFRED leaderboard. Our empirical results suggest that our novel loss objective improves generalization to unseen environments, especially by alleviating the difficulty of detecting small objects and interpreting rare words -- which are challenging error conditions in current state-of-the-art models.

\section{Proposed Approach}

\begin{figure}
    \centering
    \includegraphics[width=0.4\textwidth]{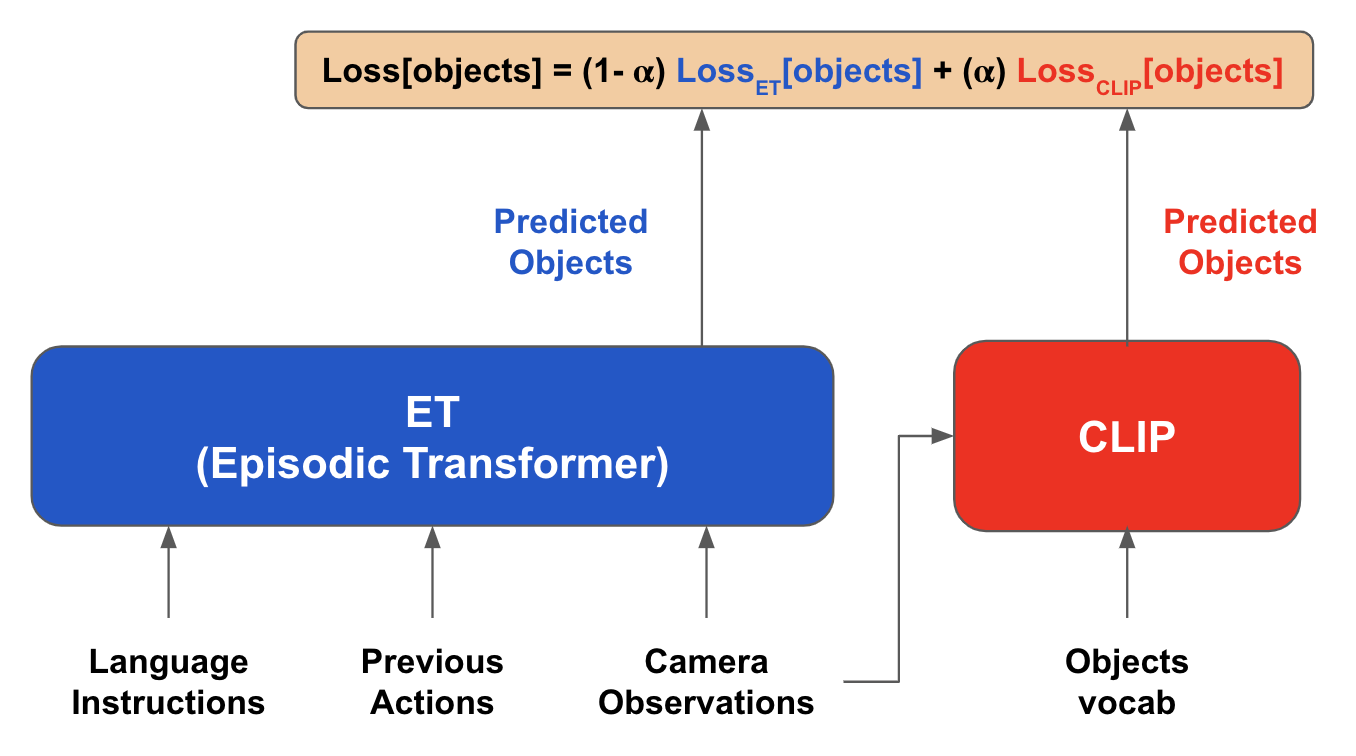}
    \caption{ET-CLIP model as modified from \cite{pashevich2021episodic}}
    \label{fig:et-clip}
\end{figure}



We use CLIP \cite{pmlr-v139-radford21a} as an auxiliary source of information for object detection and interaction by including CLIP as an additional module in ET \cite{pashevich2021episodic}. During training, we feed camera observation inputs from ET into CLIP along with a list of all ALFRED object words (with ``none" also being an option). A predicted object for each camera observation is obtained from both the CLIP module and ET, and we compute their object prediction losses: $\mathcal{L}_{\text{CLIP}}(\text{obj})$ and $\mathcal{L}_{\text{ET}}(\text{obj})$, respectively. The final object loss in ET is as follows:
\vspace{-2pt}
\begin{align*}
    \mathcal{L}{(\text{obj})} = \alpha \cdot \mathcal{L}_{\text{CLIP}}(\text{obj}) + (1-\alpha)\cdot \mathcal{L}_{\text{ET}}(\text{obj}) \\
    \text{where } \alpha \in [0,1] \quad \text{(see Figure }\ref{fig:et-clip})
\end{align*}

Both ET and CLIP weights are updated during training. During inference, the CLIP module is ignored, and object prediction is solely done by ET. 
\section{Preliminary Experiments \& Results}
\paragraph{Experimental setting} We run our baseline experiments based on the code released by the authors of the ET paper\footnote{\url{https://github.com/alexpashevich/E.T.}}. More specifically, we use the base ET model, which does not employ the data augmentation strategy. We train both the ET baseline and the ET-CLIP models for 20 epochs, and refer to the original ET model for hyperparameters. The weighting coefficient $\alpha$ of the auxiliary CLIP loss was chosen as 0.5 based on the magnitude of the two loss terms to ensure that the loss ranges are similar in the two models. 
We note that the discrepancy of our results from \cite{pashevich2021episodic} stems from different random seeds, as noted by the authors\footnote{\url{https://github.com/alexpashevich/E.T.\#et-with-human-data-only}}.


\paragraph{Results}
Table \ref{tab:experiment_results} shows the results for success rate and goal-conditioned success rate of the ET-Baseline and the ET-CLIP models for the unseen validation splits. As seen in Table \ref{tab:experiment_results}, the ET-CLIP model performs better in unseen scenes. This suggests that adding CLIP object detection as an auxiliary loss helps with generalization. We further analyze how CLIP aids in performance improvement for specific error conditions, pertaining to task instruction characteristics in Section \ref{analysis}.


\begin{table}[t!]
\centering
\begin{tabular}{l|c|c}
\toprule
Methods & {Success Rate} $\uparrow$ & {Goal-Conditioned } \\
    &   & Success Rate $\uparrow$ \\
\midrule
\hspace{2mm} ET-Baseline \cite{pashevich2021episodic}  & 0.1  & 7.8 \\
\hspace{2mm} ET-CLIP & \textbf{1.0} & \textbf{7.9} \\
\bottomrule
\end{tabular}
\caption{Success rates and goal-conditioned success rates of the baseline Episodic Transformer (ET) model and our ET-CLIP model on the unseen validation set.}
\label{tab:experiment_results}
\end{table}
\section{Analysis \label{analysis}}
We investigate how integrating CLIP helps the ET model's performance on natural language directives. In particular, we look into three subsets of instructions that contain common sources of error: instructions including fine-grained object properties, small objects, and rare semantics. We report our results in Table \ref{tab:analysis}.

\begin{table}[t!]
\centering
\begin{tabular}{@{}l|cc|c}
\toprule
 Subset & ET  & ET-CLIP & Improvement \\
\midrule
\hspace{2mm} All & 7.8 & \textbf{7.9} & + 0.1 \\
\midrule
\hspace{2mm} Object properties & 7.7 & \textbf{8.0} & + 0.3  \\
\hspace{2mm} Small objects & 5.1 & \textbf{5.6} & + 0.5 \\
\hspace{2mm} Rare semantics & 5.9 & \textbf{6.7} & + 0.8  \\
\bottomrule
\end{tabular}
\caption{Goal-conditioned success rates on the unseen validation set of the ET-Baseline and ET-CLIP on subsets of instructions.}
\label{tab:analysis}
\end{table}

\paragraph{Object properties} Interestingly, we find that ET-CLIP excels at instructions, noting specific object characteristics such as colors (e.g., ``Turn around, walk to the \textit{red} arm chair''), improving the goal-conditioned success rate by 0.3\%. The addition of our CLIP module facilitates the model to leverage specific visual cues stated in the language directives more effectively, due to the vision-language alignment learned from pre-training. This is important for correct object detection in embodied interaction tasks, especially when the environment requires semantically disambiguating objects of the same class. 

\paragraph{Small objects} Existing state-of-the-art models in ALFRED struggle with detecting small objects \cite{min2021film,pashevich2021episodic,zhang-chai-2021-hierarchical}, as they take up a negligible portion of the input image. The range of success rates in this instruction subset (5.1-5.6) is lower compared to the global average (7.8-7.9), which aligns with previous findings. Surprisingly, ET-CLIP improves the goal-conditioned success rate by 0.5\% in instructions that involve manipulating smaller objects, such as ``pencil'' or ``keys''. As the pre-trained CLIP model is trained with image-caption pairs, it is likely that the resulting representations are conducive to the semantics of the image, even when objects are small in size. 

\paragraph{Rare semantics} We additionally validate the hypothesis whether CLIP helps ET better understand instructions with rare words, which we define as words that appear less than 30 times in the training set. Since CLIP is trained with numerous captions, it is likely that ET-CLIP can benefit from this knowledge and in turn interpret rare words better than the baseline. Our results show that ET-CLIP improves ET by 0.8\% for rare semantics, which affirms our hypothesis.

\section{Conclusion}

In this work, we explore the potential of incorporating pre-trained CLIP encoders to the ALFRED task. The novelty of our method lies in leveraging CLIP as an additional module through an auxiliary object detection loss. Our approach can be easily applied to other models that employ object detectors. Our modification upon the Episodic Transformer model shows that using CLIP improves task performance especially in unseen environments, enhancing the model's ability to deal with object properties, small objects, and rare semantics. In future work, we hope to validate the effectiveness of our approach on other models in the field of embodied instruction following, to further improve where current models are failing.
\section*{Acknowledgements} 
\noindent We would like to thank Yonatan Bisk for his guidance throughout this project.

{\small
\bibliographystyle{ieee_fullname}
\bibliography{egbib}
}

\end{document}